# A Proposed IoT Smart Trap using Computer Vision for Sustainable Pest Control in Coffee Culture

Vitor A. C. Figueiredo[1], Samuel B. Mafra[1], Joel J. P. C. Rodrigues[2,3]

[1]National Institute of Telecommunications – (INATEL)
510, João de Camargo Avenue – 37.540-000 – Santa Rita do Sapucaí – MG – Brazil

[2] Federal University of Piauí (UFPI) - Ministro Petrônio Portela Avenue - 64.049-550 - Teresina - PI - Brazil.

[3] Instituto de Telecomunicações, Portugal

vitor.campos@mtel.inatel.br, samuelbmafra@inatel.br, joel@ufpi.edu.br

***Abstract.*** *The Internet of Things (IoT) is emerging as a multi-purpose technology with enormous potential for improving the quality of life in several areas. In particular, IoT has been applied in agriculture to make it more sustainable ecologically. For instance, electronic traps have the potential to perform pest control without any pesticide. In this paper, a smart trap with IoT capabilities that uses computer vision to identify the insect of interest is proposed. The solution includes 1) an embedded system with camera, GPS sensor and motor actuators; 2) an IoT middleware as database service provider, and 3) a Web application to present data by a configurable heat map. The demonstration of proposed solution is exposed and the main conclusions are the perception about pest concentration at the plantation and the viability as alternative pest control over traditional control based on pesticides.*

## 1. Introduction

It is unquestionable the importance of agriculture for humanity, since it provides foods and medical products. However, some types of insects (called pests) are undesirable in agriculture as they decrease productivity as well as cause considerable financial losses.

Taking the example of coffee culture, according to [Damon 2015], one of the most problematic pests is known as Coffee-Berry-Borer - CBB (*Hypothenemus hampei*). They are small black beetles and the adult's females measure 1.7 mm in length. The females are responsible for piercing the coffee fruit and putting their eggs. The larvae live inside the fruit, usually feeding only one seed (the fruit has two seeds) partially or totally destroying the fruit – as seen in Figure 1. As [Carvalho 2016] mentions, these beetle attacks the green fruits, mature and dry ones.

According to [Pereira 2006], [De Souza *et al*., 2016] and [Medeiros Lima 2016], the CBB in coffee-growing may causes the following damages: 1) premature loss of fruit; 2) reduction of grain weight (quantitative loss); 3) reduction of coffee beverage category (qualitative loss).

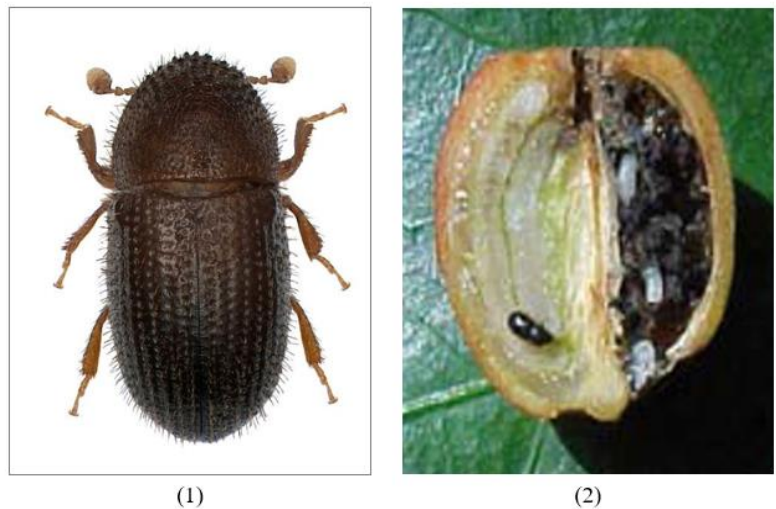

**Figure 1: The coffee-berry-borer (CBB): (1) its body shape; (2) in a coffee fruit and their larvae growing inside (right side).**

As [Carvalho 2016] describes, there are some strategies for CBB control: 1) Chemical: use of insecticides with active ingredients; 2) Homemade traps: made from plastic bottle using attractive substance formulated to coffee powder and ethanol; 3) Biological: use of fungi and insects, naturally enemies; 4) Cultural: harvest well done and eradicate abandoned plantations.

According to [Bustillo and Villalba 2004], the chemical control is the most widely applied for efficiency and practicability reasons. However, all insecticides available in market are extremely toxic and may cause health problem in humans and damage in nature life. These insecticides are also expensive and require several applications in the plantations during the year.

Based on survey published at [Martineau and Conte 2017], new solutions have been attempted motivated by the evolution of a series of technologies such IoT, cloud computing, electronic traps, and advanced insect's identification techniques. The Digital Agriculture is emerging as a trend offering technological artifacts to face the problems related to plantations on farms. [Connolly 2018] mentions that plenty of these artifacts can be combined to build an end-to-end solution and provides an innovative strategy for CBB population control. The key objective is protecting the human health and natural life. Secondary objectives are decrease economic cost pest control, the possibility to monitor the trapping operation and to receive the effectivity feedback in real-time.

The fan-based trap is inspirited in the work presented [Santos 2019]. The image-based insect identification techniques are described in [Wang 2012].

This paper is organized as follows. Section 2 presents the proposal, details the solution, and describes the software and hardware elements. Section 3 presents and analyses the obtained results. Section 4 concludes the paper and suggests future works.

## 2. Proposal and Prototyping

The Figure 2 illustrates the end-to-end solution overview that is composed by 1) the smart trap structure, including the embedded hardware (GPS sensor, batteries, relays, micro processing board); 2) an IoT middleware deployed on a cloud computing environment; 3) an Web application built over geo-analytic framework for data visualization.

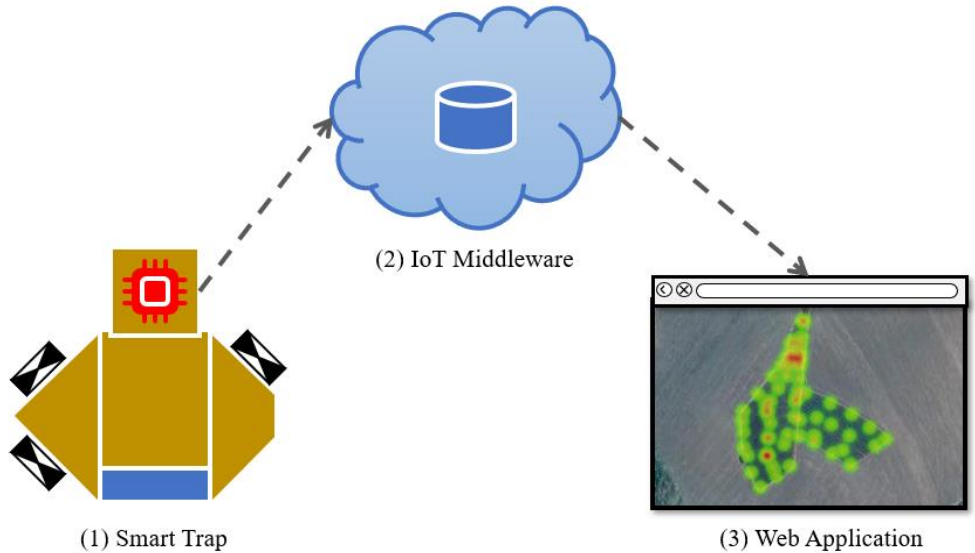

**Figure 2: End-to-end solution overview.**

Briefly, whenever a capturing event happens, the embedded software gets the geo localization (latitude and longitude), builds a message and sends it to an IoT middleware. The IoT middleware uses its internal database to store the message. On the other side, the Web application reads the data from IoT middleware and, then, end-users will be able to view the captured pest concentration through a heat map.

### 2.1. Smart Trap

Figure 3 shows the smart trap main components. The embedded hardware is composed by a Raspberry version 3 – [RPF 2020], a *Raspian Buster version September 2019* as Operation System (a Linux system flavor). The embedded hardware container also includes a GPS sensor, two-channel relays - to actuate the DC (*Direct Current*) fans, two batteries (one with 5 volts for the Raspberry and 12 volts for DC fans). The smart trap uses 1 fan for capture actions and 3 fans for eject actions.

A specific attractive substance must be used to attract the CBB. According to [Medeiros and Lima 2016], this attractive substance can be formulated by mixing methanol, ethanol, and coffee powder. In order to spread the substance smell, there is a software component responsible for actuate all Eject DC Fans every 180 seconds.

The camera is continuous taking pictures. When the beetle enters by the trap entrance, sequential pictures are taken and a software component performs, for each one, the image processing searching the CBB size body pattern. This software component extends a computer vision library.

As soon as the size body pattern is matched, the Capture DC Fan is actuated pushing the beetle against the trap cage where is trapped and killed. The trap cage is filled by water and detergent. Additionally, a software component reads the current geo localization from a GPS sensor, creates a message (with latitude, longitude, and the number of captured CBBs) and send it to an IoT Middleware. The connectivity from Raspberry Pi to the IoT Middleware is based on Wi-Fi 802.11ac standard, specified by [Wi-Fi Alliance 2020]. Otherwise (the pattern is unknown), Eject DC Fans are actuated and expels the beetle out of the trap.

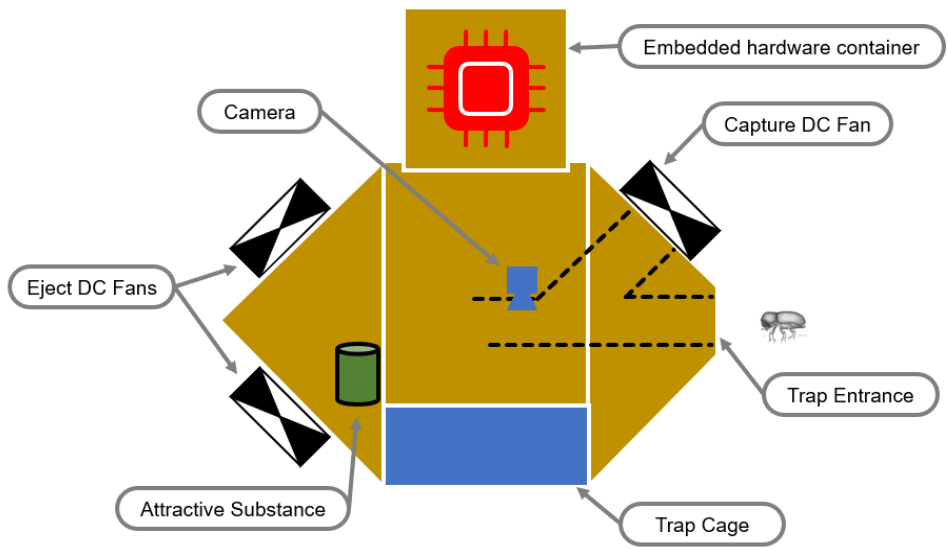

**Figure 3: Illustration of Smart Trap components.**

Figure 4 shows the photos of the created smart trap prototype. By the front view (Figure 4-a), it is possible to see the entrance, the capture DC fan, and the embedded hardware container. By the back view (Figure 4-b), the 3 DC fans for eject can be seen. Finally, the camera and the cage are seen by the internal view (Figure 4-c).

As explained by [Damon 2015], given the little capacity of flight of CBB, the trapping operation includes to move slowly the smart trap structure through coffee plantation while CBB specimens are going to be attracted and captured.

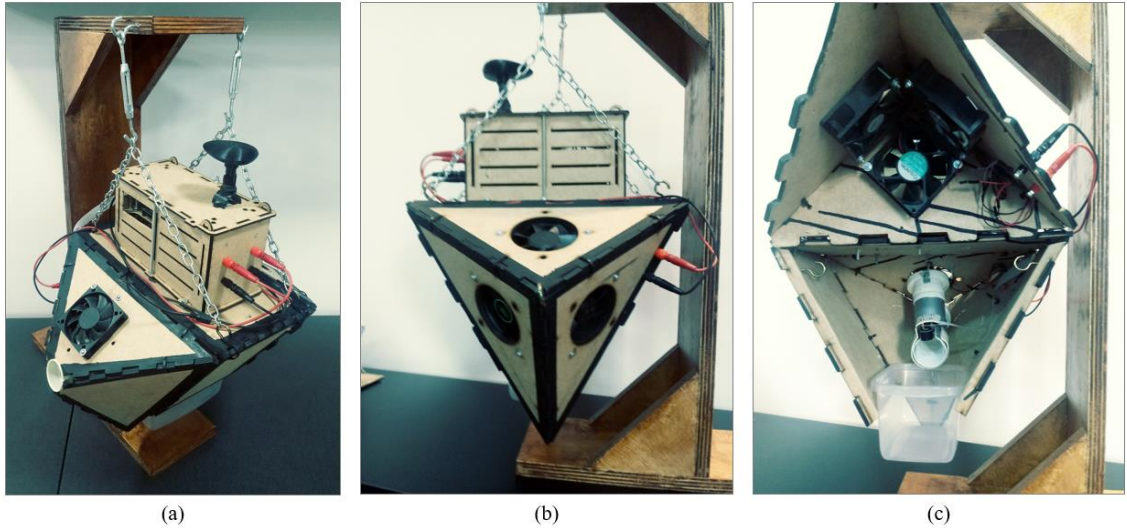

**Figure 4: Photos of the Smart trap prototype:
(a) front view; (b) back view; (c) internal view.**

Figure 5 illustrates the state machine diagram implemented by the main software component. When the state is on DETECTION MODE, the software component continually performs the image processing (detailed in Section 2.2) which result is how many CBB´s and unknown insects were found. If only CBB´s were found (one or any), the state goes to CAPTURE MODE. Otherwise (i.e. some unknown insect found), the state goes to EJECT MODE. During the CAPTURE MODE, the capture fan is actuated in order to push the CBB(s) to trap cage where it is captured and killed. Then the geo-localization is retrieved from GPS sensor, a message containing the latitude, longitude and number of CBBs is built and it is sent to IoT middleware where the message is stored. On the other hand, in EJECT MODE, all three eject fans are actuated in order to expel the unknown insect(s) (one or many).

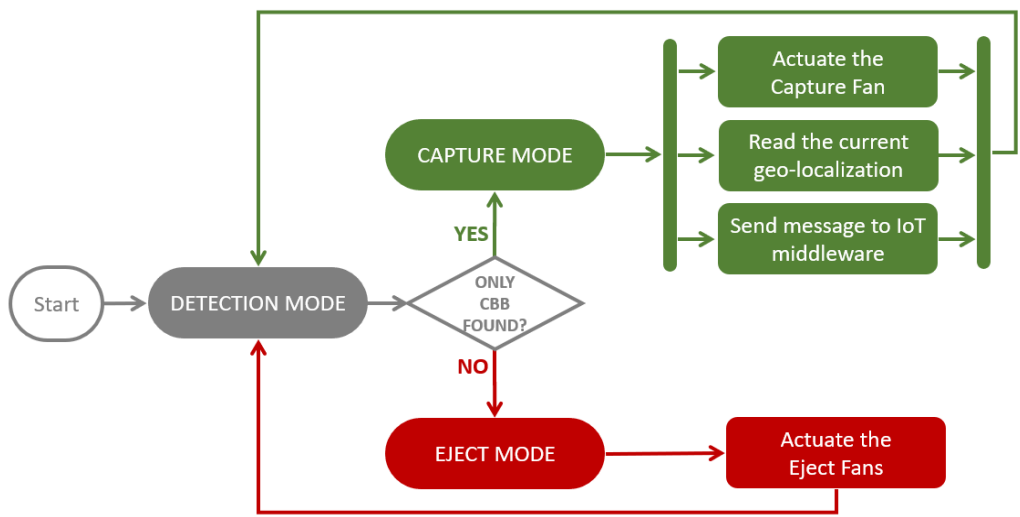

**Figure 5: State machine diagram.**

## 2.2. Image Processing

As mentioned previously, the image processing is performed during the DETECTION MODE and its goal is count how many CBBs were found as well how many unknows insects were found. The image processing considers four sub-processes, as follows:

1) Image taking

The camera is strategically positioned in front to a white background and, every second, a color photo with 640x480 resolution is taken and RGB (*Red Green Blue*) channels model. The Figure 6 (a) shows the original image taken. Then, it is converted into two in-memory pixel array: RGB and grayscale pixel arrays. Working with grayscale arrays is pre-condition for binarization process. It is worth mentioned that a color of a pixel is represented by 3 numbers (from 0 until 255) where each number defines channel intensity. For example, the notation RGB(255, 100, 50) means that the red channel (R) intensity is 255, the green channel (G) intensity is 100 and the blue channel (B) intensity is 50. For grayscale pixel representation, the 3 channel intensity are equals, where RGB(0, 0, 0) is a black pixel, RGB(127, 127, 127) is the mid-gray pixel and RGB(255, 255, 255) is the white pixel. Therefore, the grayscale has 255 possible gray colors.

2) Binarization

The binarization process consists in convert each grayscale pixel to a black or white pixel. It is important to turn the image processing simpler and faster. So, in order to performs the pixel binarization, it had to be defined one of 255 possible gray color that will be the threshold color. In other words, if a gray pixel is lighter than the threshold color, the pixel is binarized to white. And if a pixel is darker than the threshold, the pixel is binarized to black. Empirically, (considering the CBB body color), the optimal found threshold color was RGB(45, 45, 45). Once whole image is binarized, the image must be inverted, where the background (originally in white) is inverted to black area and insect body area is inverted to white area. This inversion process is useful to optimize the contour searching (explained later). The Figure 6 (b) shows the image as result after binarization and inversion. Note that some noises are present in both black and white in the image.

3) Morphological operations

The remaining noises in inverted image must be removed and, for this, two morphological operations are applied: a) the closing morphological operation eliminates small white points inside background image (black area); 2) the closing morphological operation removes black noises inside the white area. The Figure 6 (c) demonstrates the morphological operations result.

4) Contour searching

This process consists in searching for contour around the white areas (possible insect body areas). For each contour found, a rectangle is drawn around it. If its width and height are inside a predefined values range, the contour is considered the CBB and a green rectangle is drawn around it. Empirically, the defined width and height is from 10 until 60 pixels range. Otherwise, the contour is unknown and a red rectangle is drawn around it. Once all the contours are found and analyzed, the number of CBB specimens is computed as well as the number of unknown insects. Figure 6 (d) demonstrates the generated rectangles (both green and red ones) are overlaid on the original image and it is possible count the CBB and unknown insects. In this example, it shows 11 CBB samples and 3 unknown insects were found.

As already mentioned, if only CBBs were found, the state change to CAPTURE MODE, as seen in Figure 5. If some unknown insects where found (even if some CBB was found), the state changes to EJECT MODE.

The OpenCV (Open Source Computer Vision) library version 4.0 was elected to be used in the project. The main reason is its maturity, performance, and popularity within the developer community. Further information about OpenCV can be found in [OpenCV Team, 2020].

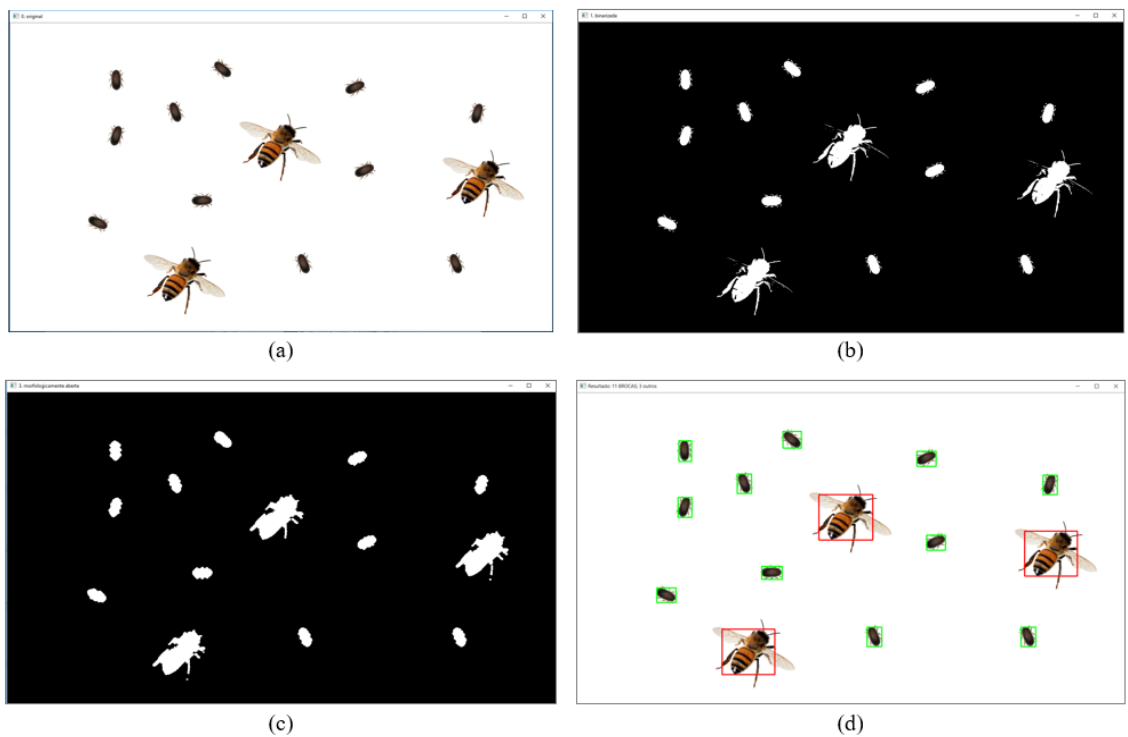

**Figure 6: Image processing samples: (a) original color image; (b) binarized image; (c) after morphological operations applied; (d) final image contoured by green rectangles (CBB detected) and red rectangles (unknown insect).**

### 2.3. IoT Middleware

An IoT middleware has two main roles: 1) it stores data pushed from the smart trap software component; and 2) it provides data to the Web application component. This project uses the In.IoT, created by [Cruz and Rodrigues 2018]. The In.IoT is a mature IoT middleware platform and it was already well used in other projects. The In.IoT platform is a state-of-the-art IoT middleware, including the needed security requirements, reporting and charting, offers a Web interface and modern communication models support.

### 2.3. Data Visualization

A Web application was developed over a geo-analytic framework. A software component reads data (containing an array of latitude, longitude, and number of captured beetles) from the IoT middleware and show it as a heat map to the end-user. The Web application also provides configurable settings like refresh screen rate, blur, and opacity adjustment and map layers (terrain, satellite, and hybrid).

The Leaflet (described in [Agafonkin 2020]) is the geo-analytic library and its plugins: 1) Leaflet.heat: for heat map support; and 2) Leaflet.GoogleMutant: for satellite images support. The Web application was developed in Java Enterprise Edition version 7 and deployed on Google Cloud Platform.

## 3. Demonstration and Validation of the Proposed Solution

The methodology for obtaining the results consisted of moving the entire trap across the plantation, stopping every 3 meters for 30 minutes. The experiments in field were performed in a coffee plantation with high infestation history. The elected farm is located in Santana da Vargem municipality, Minas Gerais State, Brazil. The Figure 7 shows the results obtained in field. The first heat map (Figure 7-a) is the closest image where each red point represents a CBB captured and its geo-location. As the zoom moves away (Figure 7-b, c, d), the red points are merged providing different perceptions about pest concentration.

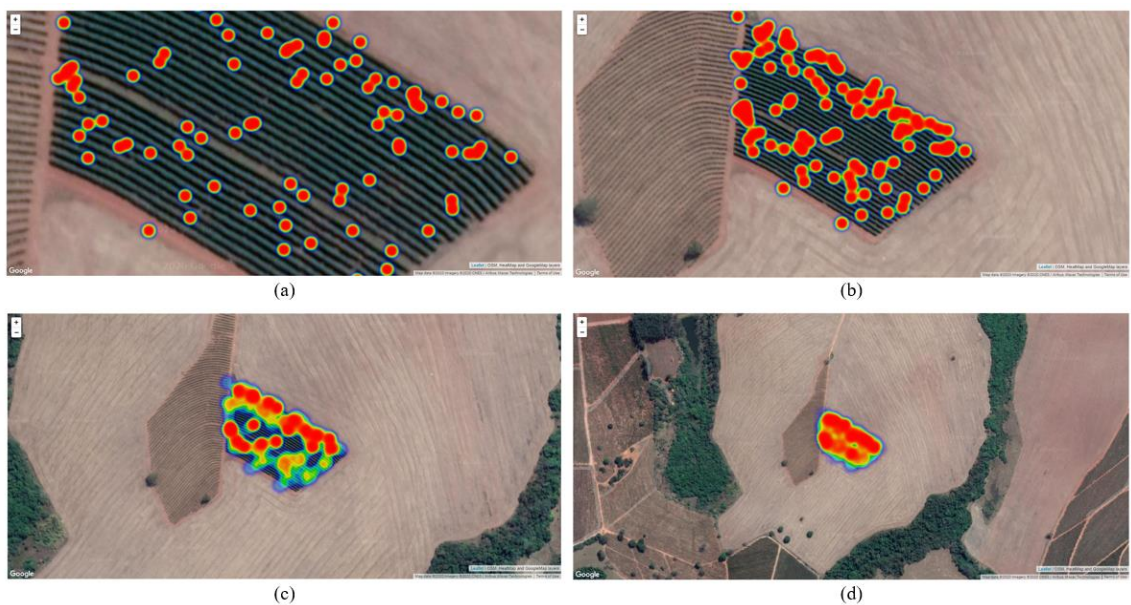

**Figure 7: Heat map superimposed on coffee plantation in different zoom levels (in Google Maps scale): (a) zoom level 16; (b) zoom level 15; (c) zoom level 14; (d) zoom level 13.**

## 4. Conclusion and Future Works

Analyzing the results, it is possible to conclude the proposed solution presents the following aspects: 1) for the particular coffee plantation, the pest concentration is uniformly distributed for whole area. To completely eliminate the pest, successive trap rounds across the plantation must be performed. At the end of every round, it is expected the pest concentration would be dispersed; 2) the heat map represents an intuitive way to view the pest concentration; 3) for the sustainable and safety reasons, it is a viable alternative pest control over traditional control based on pesticides.

Considering the present smart trap must be moved manually, many ideas have been emerged to make the process more automatic and remotely managed. Therefore, it is possible to mention some future works as follows: 1) development of an autonomous rover capable to move inside the coffee plantation guided by the geo-processing and computer vision combination. The rover might use electric motors and recharge its batteries using solar energy. The smart trap would be transported by the rover; 2) adapting the smart trap to be multi-insect. In others words, the image processing software component might be configurable to detect several types of pest and across others cultures such orange, soy, and corn; 3) alternative connectivity strategies also must be experimented. The present solution was based only on Wi-Fi connectivity. Technologies such LoRA, specified in [Lora Alliance 2020], and 5G, specified in [ITU 2019], might be suitable for farm environment.

## Acknowledgements

This work was partially supported by RNP, with resources from MCTIC, Grant No. No 01250.075413/2018-04, under the Radiocommunication Reference Center (Centro de Referência em Radiocomunicações - CRR) project of the National Institute of Telecommunications (Instituto Nacional de Telecomunicações - Inatel), Brazil; by FCT/MCTES through national funds and when applicable co-funded EU funds under the Project UIDB/EEA/50008/2020; and by the Brazilian National Council for Research and Development (CNPq) via Grants No. 431726/2018-3 and 309335/2017-5.